\documentclass[letterpaper, 10 pt, conference]{ieeeconf}

\IEEEoverridecommandlockouts
\def\BibTeX{{\rm B\kern-.05em{\sc i\kern-.025em b}\kern-.08em
    T\kern-.1667em\lower.7ex\hbox{E}\kern-.125emX}}
    
\usepackage{cite}
\usepackage{amsmath,amssymb,amsfonts}
\usepackage{algorithmic}
\usepackage{graphicx}
\usepackage{textcomp}
\usepackage{xcolor}
\usepackage{multirow}
\usepackage{caption}
\usepackage{subcaption}
\usepackage{balance}
\usepackage{url}
\usepackage{booktabs}

\usepackage{algorithm}

\usepackage{xcolor}
\usepackage{soul}
\sethlcolor{yellow}

\newcommand\MODELNAME{STT-CL}

\title{
\normalsize 2025 IEEE International Conference on Systems, Man, and Cybernetics (SMC)
\\[1em] 
\LARGE \bf
Spatial-Temporal Transformer with Curriculum Learning for EEG-Based Emotion Recognition
}

\author{
Xuetao Lin$^{1,2}$, Tianhao Peng$^{1,2}$, Peihong Dai$^{1,2}$, Yu Liang$^{3}$, Wenjun Wu$^{1,2,*}$\\
$^{1}$Beihang University, Beijing, China 
$^{2}$SKLCCSE, Beijing, China\\
$^{3}$Beijing University of Technology, Beijing, China\\
\{xtlin, pengtianhao, 20373367, wwj09315\}@buaa.edu.cn, yuliang@bjut.edu.cn
\thanks{* Wenjun Wu is the corresponding author.}
\thanks{This work is supported by State Key Laboratory of  Complex \& Critical Software Environment (SKLCCSE) and the Beijing Natural Science Foundation (No.4254091)}
\thanks{Author manuscript accepted for SMC 2025. Final version © 2025 IEEE.
}
}

\begin{document}

\maketitle
\thispagestyle{empty}
\pagestyle{empty}

\begin{abstract}
EEG-based emotion recognition plays an important role in developing adaptive brain-computer communication systems, yet faces two fundamental challenges in practical implementations: (1) effective integration of non-stationary spatial-temporal neural patterns, (2) robust adaptation to dynamic emotional intensity variations in real-world scenarios.
This paper proposes \MODELNAME{}, a novel framework integrating spatial-temporal transformers with curriculum learning. Our method introduces two core components: a spatial encoder that models inter-channel relationships and a temporal encoder that captures multi-scale dependencies through windowed attention mechanisms, enabling simultaneous extraction of spatial correlations and temporal dynamics from EEG signals.  Complementing this architecture, an intensity-aware curriculum learning strategy progressively guides training from high-intensity to low-intensity emotional states through dynamic sample scheduling based on a dual difficulty assessment. Comprehensive experiments on three benchmark datasets demonstrate state-of-the-art performance across various emotional intensity levels, with ablation studies confirming the necessity of both architectural components and the curriculum learning mechanism.
\end{abstract}

\section{Introduction}
Emotion recognition constitutes a fundamental component of brain-inspired human-computer interaction systems~\cite{duan2023gender}. 
Electroencephalography (EEG), a non-invasive technique for measuring brain activity with high temporal resolution, has emerged as a critical modality for emotion recognition due to its ability to directly capture neural dynamics objectively~\cite{liang2024fetcheeg}. Compared to non-physiological modalities such as facial expressions and speech, EEG provides a more direct reflection of the underlying cognitive and emotional processes, making it a preferred way for fine-grained emotion recognition~\cite{li2023eeg}.

EEG-based emotion recognition is challenging due to the non-stationary, noisy, and high-dimensional nature of EEG signals~\cite{wang2024research,peng2025goat}. Recent advancements in deep learning have significantly improved EEG-based emotion recognition by enabling more efficient modeling of spatial and temporal dependencies in EEG features~\cite{liang2022help,peng2024pre}.
Early methods like convolutional neural networks (CNNs)~\cite{lawhern2018eegnet} and graph-based approaches~\cite{peng2023clgt, peng2024graphrare,peng2025sola, chen2022microegrcl} such as DGCNN~\cite{song2020eeg} have proven effective in extracting spatial features, while models like ATDD-LSTM~\cite{du2022efficient} and Transformer-based architectures~\cite{wang2022transformers, song2023eeg} focus on capturing temporal dependencies~\cite{peng2025tagrec}. However, these methods often treat spatial and temporal features independently, limiting their ability to fully model the joint spatiotemporal dynamics critical for emotion recognition. 

In addition to the spatial and temporal complexities of EEG signals, emotional intensity variability presents a significant challenge in emotion recognition. Emotional intensity, defined as the strength of an emotional state, strongly influences the neural patterns observed in EEG signals~\cite{gross1998emerging}. High-intensity emotions are often easier to classify due to their distinct neural activations, whereas low-intensity emotions produce subtle patterns that are more challenging to detect~\cite{russell1980circumplex}. Most existing frameworks assume that emotional states remain stable during experimental trials, failing to account for the dynamic fluctuations in intensity observed in real-world scenarios~\cite{can2023approaches}.
Several approaches have been proposed to address emotional intensity variability. These include using generative models like Variational Autoencoders (VAEs)~\cite{tian2023dualencoder, behrouzi2022graph} and Generative Adversarial Networks (GANs)~\cite{zhang2021ganser, pan2021emotion} to augment datasets, and employing contrastive learning to align feature representations across intensity levels~\cite{shen2024contrastive}. While effective for data imbalance and representation learning, these methods lack a structured, progressive mechanism for capturing the hierarchical nature of emotional intensity variations.

In summary, despite recent advancements in deep learning and EEG-based emotion recognition, two critical challenges remain unresolved:

(1) EEG signals exhibit complex temporal and spatial dependencies that are difficult to model jointly. While existing methods focus on either spatial (e.g. CNNs, GNNs) or temporal features (e.g. LSTMs, Transformers), they often fail to capture the full spatiotemporal dynamics essential for understanding emotional states comprehensively. 

(2) Emotional intensity fluctuates over time, most existing models assume emotional stability and lack mechanisms to adapt to these intensity variations, resulting in reduced robustness in real-world scenarios.

To address these challenges, we propose the Spatial-Temporal Transformer with Curriculum Learning (\MODELNAME{}) for EEG-Based Emotion Recognition, a novel framework designed to capture the spatiotemporal dynamics of EEG signals while systematically addressing emotional intensity variability. 

The contributions of \MODELNAME{} can be summarized as follows: \begin{enumerate} 
    \item We introduces a dual-attention mechanism to simultaneously model inter-channel spatial relationships and long-range temporal dependencies, addressing the limitations of existing Transformer-based models.
    \item The proposed \MODELNAME{} implements a curriculum learning framework that dynamically adapts to emotional intensity strata through sample difficulty scheduling, enhancing classification robustness intensity levels.
    \item Extensive experiments on EEG datasets demonstrate the effectiveness of \MODELNAME{}, showcasing its ability to handle emotional intensity variability and capture spatial-temporal dynamics. 
\end{enumerate}

    


\section{Related Work}

\subsection{Spatial Temporal Modeling Methods}
EEGNet~\cite{lawhern2018eegnet} employs depthwise and separable convolutions to extract spatial and temporal features efficiently but struggles to model long-range dependencies critical for complex emotional states. FBCCNN~\cite{pan2021emotion} and Tsception~\cite{ding2023tsception} improve spatial and temporal feature extraction, respectively, but fail to jointly model spatiotemporal relationships in EEG data. Graph-based methods like DGCNN~\cite{song2020eeg} and RGNN~\cite{zhong2022eegbased} capture inter-channel spatial correlations by modeling EEG signals as graphs. However, these methods lack explicit mechanisms to account for temporal dynamics. Hybrid models, such as ECLGCNN~\cite{yin2021eega}, combine graph-based spatial modeling with LSTMs for temporal feature extraction but rely on predefined adjacency matrices limiting adaptability to dynamic EEG patterns. Transformer-based architectures have emerged as promising alternatives for modeling EEG data. HSLT~\cite{wang2022transformers} extracts hierarchical spatial features, while EEG Conformer~\cite{song2023eeg} combines convolutional layers for local spatial patterns with Transformer layers for long-range temporal modeling. However, these models often treat spatial and temporal features separately, limiting their ability to fully capture the joint spatiotemporal dynamics essential for emotion recognition.

\subsection{Emotional Intensity Modeling Methods}
Emotional intensity variability poses a significant challenge in EEG-based emotion recognition. To address this, feature augmentation techniques such as VAE~\cite{tian2023dualencoder, behrouzi2022graph} and GAN~\cite{pan2021emotion, zhang2022ganser} generate synthetic low-intensity samples, improving dataset balance and classifier robustness. However, these methods are computationally expensive and sensitive to hyperparameter tuning, limiting their scalability~\cite{he2021data}. Contrastive learning approaches, such as the spatiotemporal framework by Shen et al.~\cite{shen2024contrastive}, align feature representations across intensity levels to enhance generalization. Yet, they lack hierarchical learning mechanisms to handle the progressive complexity of emotional intensity variations. 

Curriculum learning~\cite{bengio2009curriculum} provides a systematic solution that plan the training process of models in various domains~\cite{yu2024adaptaug,yu2024leveraging,feng2024safe,feng2024hierarchical}. Yang et al.~\cite{yang2022hybrid} applied CL to text-based emotion recognition by prioritizing high-confidence samples, improving model stability. Zoumpourlis et al.~\cite{zoumpourlis2024motor} used CL to enhance cross-subject generalization in motor imagery decoding by progressively increasing task complexity. Feng et al.~\cite{feng2023semantic} extended CL to EEG-to-text generation, introducing ambiguous features progressively. Inspired by these works, we adapt CL to EEG-based emotion recognition, leveraging emotional intensity as a natural criterion to guide training and effectively learn subtle low-intensity patterns.

\section{Motivation}
In EEG-based emotion recognition, current experimental paradigms assign identical emotional labels to all EEG segments extracted from the same stimulus-evoked trial, implicitly assuming uniform emotional intensity levels across co-labeled segments. This fundamental assumption inherently contradicts real-world neurodynamics where affective intensity fluctuates temporally, even within single trials. To bridge this experimental-neurophysiological gap, we implement curriculum learning with dynamic sample scheduling guided by evolving intensity estimates. Specially, we operationalize emotional intensity through a neurocomputational lens as sample-wise discrimination difficulty, quantified by a composite metric combining instantaneous prediction loss and cumulative error history. This dual-aspect quantification naturally captures both transient neural activation patterns and persistent learning challenges inherent to low-intensity states. The framework progressively exposes the model to graded intensity strata—from prototypical high-discriminability to ambiguous low-intensity patterns—through neural-plausible adaptation mechanisms. This strategic progression enhances robustness to affective fluctuations while maintaining methodological rigor through data-driven intensity quantification.




\begin{figure*}[]
	\centering 
    \includegraphics[width=0.75\linewidth]{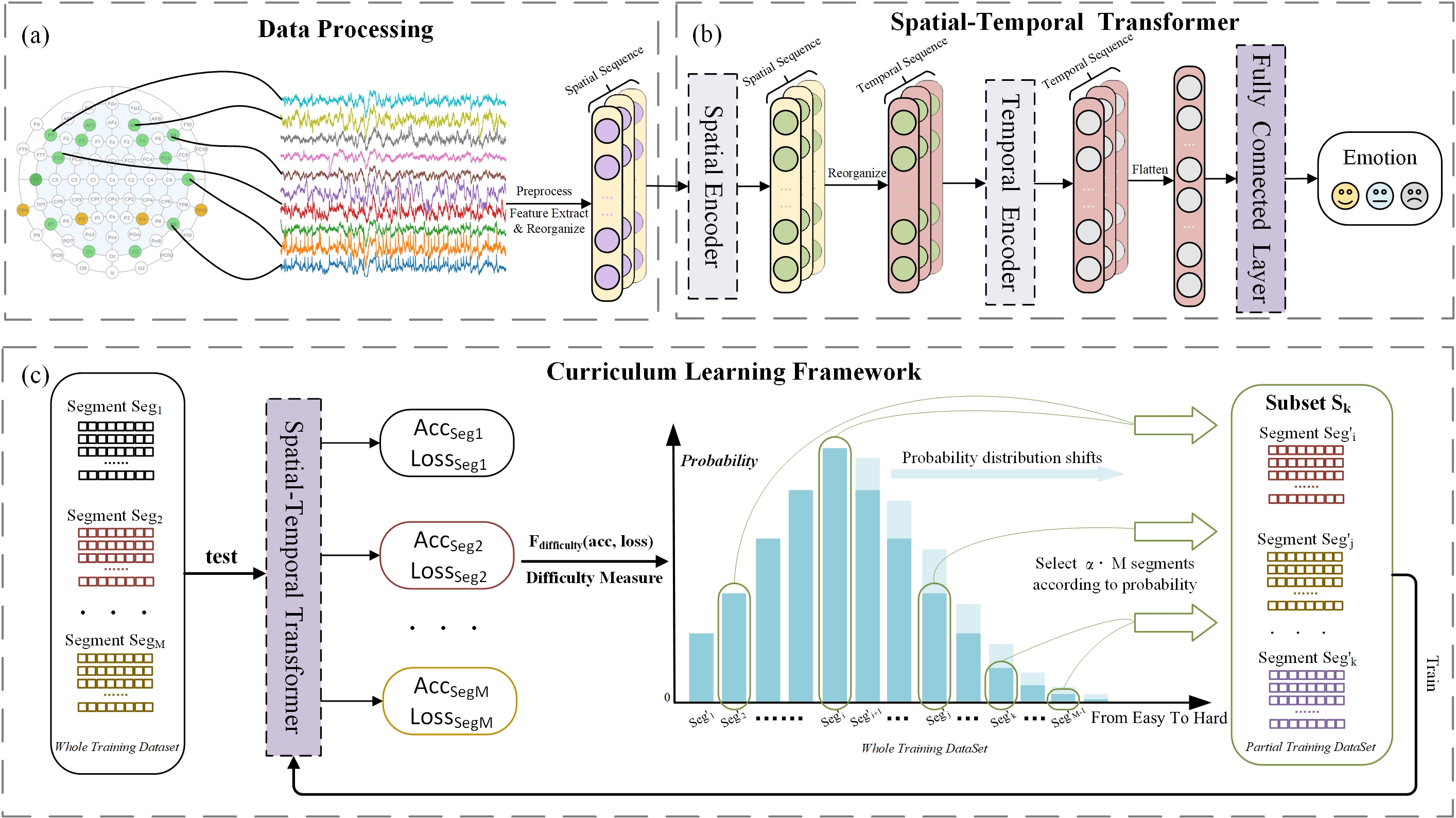} 
    \caption{Overview of the proposed \MODELNAME{}. (a) Raw EEG signals undergo preprocessing, feature extraction, and reorganization to construct a spatial-series format.(b) The processed EEG data is first fed into the Spatial Encoder, then reorganized into a temporal sequence before passing through the Temporal Encoder. (c) is an illustration of our curriculum learning process, where training samples are dynamically organized from "easy" to "hard" based on their difficulty levels. As training progresses, probability density function evolves from left-skewed (prioritizing easy samples) to right-skewed (emphasizing hard samples) to progressively handle increasingly complex emotional states. Here, $\text{Acc}_{\text{Seg}_i}$ denotes the cumulative accuracy of segment $i$, and $\text{Loss}_{\text{Seg}_i}$ represents the cross-entropy loss of sample $i$.}
\label{fig:framework}
\end{figure*}

\section{Methodology}
In this section, we present the proposed \MODELNAME{} framework for EEG-based emotion recognition. \MODELNAME{} is designed to address two major challenges: (1) joint modeling of spatial and temporal dependencies in EEG signals, and (2) handling the variability in emotional intensity through progressive learning. As shown in Fig.~\ref{fig:framework}, the framework consists of three major components: (a) the data processing, (b) the Spatial-Temporal Transformer, and (c) the curriculum learning training framework.

\subsection{Data Processing}
EEG signals are recorded using electrodes placed on the scalp, 
each recording is typically represented as a matrix $\mathbf{X} \in \mathbb{R}^{C \times T}$, where $C$ is the number of channels, and $T$ is the number of time steps. However, EEG signals are noisy and high-dimensional, requiring robust preprocessing and feature extraction to improve downstream model performance.

Our processing pipeline follows prior work~\cite{zheng2015investigating} and begins with spatial normalization through common average re-referencing to mitigate channel-specific biases. Next, a Butterworth bandpass filter (0.5-48 Hz) is applied for spectral purification, suppressing low-frequency drifts (\textless 0.5 Hz) and high-frequency muscle artifacts (\textgreater 48 Hz). Independent Component Analysis (ICA), with automated ocular artifact detection~\cite{pion2019iclabel}, is subsequently used to isolate and remove physiological contaminants while preserving neural oscillatory patterns. This cascade processing ensures signal integrity while maintaining the temporal resolution essential for emotion-related neural dynamics.

After preprocessing, the continuous EEG signal is divided into continuous windows, allowing for the capture of distinct temporal segments while retaining important contextual stability across the signal. Each window undergoes spectral decomposition via Welch's method to compute Differential Entropy (DE) features across five frequency bands: $\delta$ (0.5-4 Hz), $\theta$ (4-8 Hz), $\alpha$ (8-13 Hz), $\beta$ (13-30 Hz), and $\gamma$ (30-48 Hz). The DE values are calculated as:
\begin{equation}
DE = \frac{1}{2}\ln(2\pi e\sigma^2)
\end{equation}
where $\sigma^2$ represents signal variance within each sub-band. The resulting feature matrix for each segment is denoted as $\mathbf{Seg} \in \mathbb{R}^{\mathcal{T} \times C \times B}$, where $\mathcal{T}$ represents the number of time windows in one segment, and $B$ denotes the number of frequency bands.

\subsection{Spatial-Temporal Transformer}
The Spatial-Temporal Transformer is the core component of \MODELNAME{}, designed to extract joint spatial and temporal features from EEG signals. STT consists of two key modules: the Spatial Encoder and the Temporal Encoder. The Spatial Encoder processes the spatial correlations across channels, while the Temporal Encoder captures the temporal dependencies within each EEG channel.

\subsubsection{Spatial Encoder}
The Spatial Encoder models inter-channel dependencies by analyzing the dynamics across electrodes. Giving the input $\mathbf{Seg} \in \mathbb{R}^{\mathcal{T} \times C \times B}$, we first reorganize it into $\mathbf{X}_{\text{Spatial}} \in \mathbb{R}^{C \times (T \cdot 
 B)}$, then pass it through the Spatial Encoder. The encoder applies multi-head self-attention to discover functional connectivity patterns:
\begin{equation}
\text{Attention}(\mathbf{Q}, \mathbf{K}, \mathbf{V}) = \text{Softmax}\left(\frac{\mathbf{Q}\mathbf{K}^\top}{\sqrt{d_c}}\right)\mathbf{V}
\end{equation}
\noindent where $\mathbf{Q} = \mathbf{X}_{\text{Spatial}}\mathbf{W}_q$, $\mathbf{K} = \mathbf{X}_{\text{Spatial}}\mathbf{W}_k$, and $\mathbf{V} = \mathbf{X}_{\text{Spatial}}\mathbf{W}_v$ are learnable projections. This attention mechanism automatically identifies salient spatial correlations through end-to-end training, eliminating the need for predefined neurophysiological constraints.

Then, the processed features undergo refinement through:
\begin{equation}
\mathbf{H}_s = \text{LayerNorm}\left(\mathbf{X}_{\text{Spatial}} + \text{FFN}(\text{Attention}(\mathbf{Q}, \mathbf{K}, \mathbf{V}))\right)
\end{equation}
yielding compact spatial representations $\mathbf{H}_s \in \mathbb{R}^{C \times (\mathcal{T} \cdot d_c)}$ that capture inter-channel relationships and preserve relevant temporal information.

\subsubsection{Temporal Encoder}
Building upon the spatial representations, the Temporal Encoder captures multi-scale temporal dependencies through a windowed attention mechanism. Given the spatial embeddings $\mathbf{H}_s \in \mathbb{R}^{C \times (\mathcal{T} \cdot d_c)}$, we reorganize it to obtain temporal sequences $\mathbf{X}_t \in \mathbb{R}^{T \times (C \cdot d_c)}$. The hybrid attention architecture enables simultaneous modeling of local and global temporal dynamics:

\begin{equation}
\mathbf{A}_t = \text{Softmax}\left(\frac{\mathbf{Q}'\mathbf{K}'^\top}{\sqrt{d_t}} \odot \mathbf{M}_{\text{win}}\right)
\end{equation}

\noindent where $\mathbf{M}_{\text{win}} \in \{0,1\}^{T \times T}$ is a sliding window to prioritize proximal time steps while retaining capacity for long-range interactions. This design captures both transient neural oscillations and sustained temporal patterns inherent in emotional responses.

The processed features are then transformed through:

\begin{equation}
\mathbf{H}_t = \text{LayerNorm}\left(\mathbf{X}_{\text{t}} + \text{FFN}(\mathbf{A}_t\mathbf{V}')\right)
\end{equation}

\noindent yielding embeddings $\mathbf{H}_t \in \mathbb{R}^{T \times d_t}$ that capture both spatial and temporal patterns across the EEG signal, as they are derived from the Spatial Encoder and the Temporal Encoder.

\subsubsection{Classification and Output}
The spatial-temporal embeddings $\mathbf{H}_t$ undergo hierarchical transformation through a cascaded classifier architecture to generate emotion probability distributions. The computational process is formally defined as:

\begin{equation}
\mathbf{y} = \text{Softmax}\left(\mathbf{W}_2 \cdot \text{GELU}\left(\mathbf{W}_1 \cdot \text{Flatten}(\mathbf{H}_t)\right)\right)
\end{equation}

\noindent where the $\text{Flatten}(\cdot)$ operation reshapes the spatial-temporal embedding tensor $\mathbf{H}_t \in \mathbb{R}^{C \times d_t}$ into a vectorized representation $\mathbf{h}_{\text{flat}} \in \mathbb{R}^{C \cdot d_t}$, preserving spatial-temporal interactions while enabling subsequent linear projections. The weight matrix $\mathbf{W}_1 \in \mathbb{R}^{(C \cdot d_t) \times d_h}$ implements a non-linear mapping through Gaussian Error Linear Unit (GELU) activation, projecting the flattened features into a latent semantic space of dimension $d_h$. Finally, $\mathbf{W}_2 \in \mathbb{R}^{d_h \times K}$ establishes discriminative boundaries in this latent space to generate class logits for $K$ emotion categories, which are normalized through softmax activation to obtain probability distributions.

\subsection{Curriculum Learning Framework}
To handle the variability of emotional intensity, \MODELNAME{} uses a curriculum learning framework that dynamically adjusts the training process based on sample difficulty. The CL framework consists of two key components: difficulty evaluation and dynamic training subset selection.The detailed process of curriculum learning is summarized in Algorithm ~\ref{alg:cl}.

\begin{algorithm}[h!]
\caption{Curriculum Learning Framework}
\begin{algorithmic}[1]
\REQUIRE Training dataset $X \in \mathbb{R}^{M \times \mathcal{T} \times C \times B}$, corresponding labels $y \in \mathbb{R}^{M}$, total number of epochs $total\_epochs$;
\ENSURE Updated model parameters;

\STATE Initialize the history accuracy table $\mathcal{H} \in \mathbb{R}^{M}$, where each element is an empty list;
\STATE Initialize the difficulty table $\mathcal{D} \in \mathbb{R}^{M}$ and the selection probability table $\mathcal{P} \in \mathbb{R}^{M}$, with all values set to 0;
\STATE Initialize the current training dataset\\
$Sample\_train \gets X$, $y\_train \gets y$;

\FOR {epoch = 1 to $total\_epochs$}
    \FOR {$(\text{sample}, y_i)$ in $(Sample\_train, y\_train)$}
        \STATE $\hat{y} = \text{model}(\text{sample})$;
        \STATE Update $\mathcal{H}[i] \gets \mathcal{H}[i] \cup \mathbb{I}(\hat{y} == y_i)$;
        \STATE Perform model training and backpropagation;
    \ENDFOR

    \FOR {$(\text{sample}', {y_i}')$ in $(X,y)$}
        \STATE $\hat{y}' = \text{model}(\text{sample}')$;
        \STATE $\mathcal{L}_i^{(k)} = \text{loss}(\hat{y}'_i, {y_i}')$;
        \STATE Update $\mathcal{D}[i] \gets d_i$ based on \eqref{eq:difficulty};
    \ENDFOR

    \FOR {$\text{sample}''$ in $X$}
        \STATE Update $\mathcal{P}[i] \gets P_i^{(k)}$ based on \eqref{eq:selection_probability};
    \ENDFOR

    \STATE Update $Sample\_train$ \text{by selecting} $\alpha \times M$ samples from $X$ based on $\mathcal{P}$;

    \STATE Sort $Sample\_train$ according to $\mathcal{D}$;
\ENDFOR

\STATE \textbf{Return} Updated model parameters
\end{algorithmic}
\label{alg:cl}
\end{algorithm}

\subsubsection{Difficulty Evaluation}

The difficulty of each EEG segment is estimated through a composite metric combining both instantaneous loss and historical performance. This evaluation is conducted over the entire training dataset at each epoch for two purposes: 1) it captures global learning patterns beyond current training subsets, thereby mitigating selection bias inherent in partial dataset assessments. and 2) maintains temporal continuity in difficulty estimation across all samples. For sample $i$ at training stage $k$, we compute:

\begin{equation}
\label{eq:difficulty}
d_i^{(k)} = \underbrace{\mathcal{L}_i^{(k)}}_{\text{Current Loss}} + \beta \cdot \underbrace{\left(1 - \frac{\sum_{t=1}^k \mathbb{I}(y_i = \hat{y}_i^{(t)})}{k}\right)}_{\text{Historical Error Rate}}
\end{equation}

\noindent where $\mathcal{L}_i^{(k)}$ is the instantaneous cross-entropy loss at current epoch $k$, $\mathbb{I}(\cdot)$ is the indicator function, and $\beta$ balances the two components. This dual-metric approach captures two temporal dimensions of learning dynamics: the current loss term directly reflects the sample's instantaneous prediction difficulty, while the error rate term evaluates long-term classification patterns through accumulated historical accuracy statistics. The parameter $\beta$ acts as an adaptive gatekeeper—lower values prioritize recent performance fluctuations for rapid adaptation, whereas higher values emphasize systematic learning bottlenecks. By integrating both temporal granularity and longitudinal consistency, the metric robustly identifies samples that require sustained learning effort, avoiding overemphasis on ephemeral anomalies while progressively exposing the model to progressively complex emotional states.

\subsubsection{Dynamic Training Subset Selection}
The sample selection process employs a shifting probability distribution to implement curriculum-driven learning.The core idea consists of two synergistically coordinated strategies:

The peak migration operator governs sample selection probabilities through an adaptive Gaussian kernel:
\begin{equation}
\label{eq:selection_probability}
P_i^{(k)} = \frac{\exp\left(-\frac{(d_i - \mu_k)^2}{2\sigma_k^2}\right)}{\sum_{j=1}^M \exp\left(-\frac{(d_j - \mu_k)^2}{2\sigma_k^2}\right)}
\end{equation}
\noindent where the dynamic center parameter $\mu_k$ controls the difficulty focus window, and the bandwidth parameter $\sigma_k$ regulates sample diversity. During early training phases, $\mu_k$ is initialized to emphasize prototypical samples with easy samples while $\sigma_k$ maintains wide exploration bandwidth. As training progresses, $\mu_k$ systematically shifts toward higher difficulty regions while $\sigma_k$ undergoes controlled narrowing, creating a smooth curriculum transition from high-intensity focus on discriminative patterns, through moderate exploration of ambiguous samples, to low-intensity refinement of subtle emotional expressions.

Complementing this spatial adaptation, the subset size scheduler regulates data exposure intensity through monotonic linear expansion:
\begin{equation}
|S_k| = \lfloor \alpha_k M \rfloor, \quad \alpha_k \in (0,1]
\end{equation}
The expansion coefficient $\alpha_k$ follows a linear progression from partial to complete dataset utilization, gradually enlarging the training subset throughout the learning process. This controlled scaling mechanism ensures continuous difficulty escalation by systematically introducing more challenging samples proportionally to the model's growing competency. Following probabilistic selection, the chosen subset $S_k$ undergoes difficulty-based sorting where samples are arranged in ascending order of their $d_i$ values, establishing explicit curriculum sequencing within each training epoch.

The compound effect of these mechanisms emulates expert-guided learning processes in neural systems - initially building strong priors through focused repetition, then refining discrimination capabilities via controlled challenge escalation. This computational curriculum mirrors the natural progression of human skill acquisition, where foundational competence in core patterns precedes sophisticated handling of edge cases and subtle variations through progressive complexity exposure.

\section{Experiments}
This section presents the evaluation of \MODELNAME{} on EEG-based emotion recognition tasks using three benchmark datasets. We first describe the datasets, baseline models, and experimental setup. Following this, we compare the performance of \MODELNAME{} against state-of-the-art methods and provide detailed analyses.

\subsection{Datasets}
We evaluate the proposed \MODELNAME{} on three widely used EEG-based emotion recognition datasets:

SEED Series (SEED~\cite{zheng2015investigating} and SEED-IV~\cite{zheng2018emotionmeter}) : The SEED dataset consists of EEG signals from 15 subjects, each participating in three sessions with 15 emotion-inducing trials per session (positive, neutral, and negative). SEED-IV extends this dataset with four emotion labels (happiness, sadness, fear, and neutral) and 24 trials per session. Both datasets were recorded with 62 EEG channels.

DEAP~\cite{koelstra2011deap}: This dataset contains EEG signals from 32 participants while they watched 40 one-minute music videos, rated on arousal, valence, dominance, and liking. EEG signals were recorded using 32 EEG channels and downsampled to 128 Hz for analysis.

\subsection{Compared Methods}
This section introduces the methods used for comparison in
our study, including spatial-temporal modeling and emotional intensity modeling baselines.

Spatial-Temporal modeling baselines:
CNN-based methods include EEGNet~\cite{lawhern2018eegnet} with depthwise convolutions and Tsception~\cite{ding2023tsception} using multi-scale temporal kernels. Graph-based approaches contain DGCNN~\cite{song2020eeg} for dynamic connectivity learning and RGNN~\cite{zhong2022eegbased} with neurophysiological constraints. Transformer architectures feature HSLT~\cite{wang2022transformers} employing hierarchical attention and EEG Conformer~\cite{song2023eeg} combining CNNs with transformers.

Emotional intensity modeling baselines:
Generative approaches comprise FBCCNN~\cite{pan2021emotion} synthesizing low-intensity samples via GANs and STNet~\cite{zhang2022ganser} utilizing adversarial segment interpolation. For curriculum learning, we implement three paradigms under Bengio's framework~\cite{bengio2009curriculum}: 1) Self-Paced CL~\cite{kumar2010self, jiang2015self} prioritizes high-confidence samples, 2) Competence-CL~\cite{platanios2019competence} schedules training via model competency metrics, and 3) Dynamic-CL~\cite{zhou2020curriculum} dynamically adjusts sample probabilities based on instance hardness.

\subsection{Experimental Setting} 
We follow the experimental protocols from LibEER~\cite{liu2024libeer} and TorchEEG~\cite{zhang2024torcheegemo}, employing a 5-fold cross-validation scheme with the cross-trial setting. In this setup, each subject’s trials are divided into five folds, with training and testing conducted on different trials, ensuring effective evaluation of the model’s performance across distinct emotional states. For SEED and SEED-IV, we perform three-class and four-class classification respectively according to their labels, while for DEAP, we derive four classes by dichotomizing arousal and valence into high and low levels. We use classification accuracy and F1-score as the evaluation metrics.

\begin{table*}[htb!]
\caption{Performance comparison on SEED, SEED-IV, and DEAP datasets. Best results are in \textbf{bold}, second-best are \underline{underlined}.}
\label{tab:performance}
\centering
\resizebox{\textwidth}{!}{%
\begin{tabular}{llcccccc}
\toprule
\multicolumn{2}{c}{\multirow{2}{*}{Model}} & \multicolumn{2}{c}{SEED} & \multicolumn{2}{c}{SEED-IV} & \multicolumn{2}{c}{DEAP} \\
\cmidrule(lr){3-4} \cmidrule(lr){5-6} \cmidrule(lr){7-8}
\multicolumn{2}{c}{} & Accuracy & F1-Score & Accuracy & F1-Score & Accuracy & F1-Score \\
\midrule

\multirow{2}{*}{\textit{CNN-based}} & EEGNet~\cite{lawhern2018eegnet} & 0.5747±0.1524** & 0.3350±0.0933** & 0.4573±0.1226** & 0.3775±0.1205** & 0.4187±0.0755** & 0.3232±0.0491** \\
& Tsception~\cite{ding2023tsception} & 0.6282±0.1340** & 0.3802±0.0845** & 0.5409±0.1243** & 0.4845±0.1191** & 0.4370±0.0675** & 0.3357±0.0557** \\
\cmidrule(lr){1-8}

\multirow{3}{*}{\textit{GNN-based}} & DGCNN~\cite{song2020eeg} & 0.8125±0.1284 & 0.7361±0.1403* & 0.7234±0.1260* & 0.6765±0.1342 & 0.5121±0.0762 & 0.4059±0.0738 \\
& RGNN~\cite{zhong2022eegbased} & 0.6315±0.1523** & 0.5148±0.1769** & 0.5679±0.1077** & 0.4464±0.1109** & 0.4806±0.0714** & 0.2815±0.0728** \\
& GCBNet~\cite{zhang2022gcbnet} & 0.7991±0.1305 & 0.6848±0.1706** & 0.7464±0.1236 & \underline{0.6985±0.1289} & 0.5089±0.0876 & 0.4092±0.0831 \\
\cmidrule(lr){1-8}

\multirow{2}{*}{\textit{Transformer-based}} & HSLT~\cite{wang2022transformers} & 0.6834±0.1658** & 0.5578±0.1793** & 0.5174±0.1071** & 0.4174±0.1133** & 0.4837±0.0841* & 0.3586±0.0783** \\
& Conformer~\cite{song2023eeg} & 0.4615±0.1476** & 0.2705±0.0684** & 0.3906±0.0826** & 0.3176±0.0832** & 0.4252±0.0816** & 0.3222±0.0576** \\
\cmidrule(lr){1-8}

\multirow{2}{*}{\textit{Generative-based}} & STNet~\cite{zhang2022ganser} & 0.4943±0.0772** & 0.3099±0.0554** & 0.3649±0.0431** & 0.2818±0.0366** & 0.4004±0.0641** & 0.3128±0.0409** \\
& FBCCNN~\cite{pan2021emotion} & 0.8163±0.1123 & 0.7366±0.1598* & 0.7032±0.1269** & 0.6518±0.1307** & 0.4960±0.0803* & 0.3920±0.0709* \\
\cmidrule(lr){1-8}

& STT(Base) & 0.8031±0.1231 & 0.6322±0.1888** & 0.7386±0.1020 & 0.6920±0.1091 & 0.4991±0.0768* & 0.3954±0.0651 \\
\cmidrule(lr){1-8}

\multirow{3}{*}{\textit{CL-based}} & STT-SPCL~\cite{kumar2010self} & 0.8107±0.1001* & 0.6970±0.1414** & 0.7425±0.1114 & 0.6901±0.1260 & 0.5036±0.0792 & 0.3917±0.0707* \\
& STT-CCL~\cite{platanios2019competence} & \underline{0.8365±0.1077} & 0.6919±0.1760** & 0.7359±0.1413 & 0.6937±0.1615 & 0.5160±0.0786 & 0.3950±0.0755 \\
& STT-DCL~\cite{zhou2020curriculum} & 0.8302±0.0642 & \underline{0.7599±0.0965} & \underline{0.7482±0.1189} & 0.6967±0.1296 & \underline{0.5258±0.0849} & \underline{0.4158±0.0824} \\
\cmidrule(lr){1-8}

& \MODELNAME{} (Ours) & \textbf{0.8384±0.0695} & \textbf{0.7824±0.1017} & \textbf{0.7652±0.1071} & \textbf{0.7172±0.1200} & \textbf{0.5299±0.0731} & \textbf{0.4226±0.0613} \\
\bottomrule
\end{tabular}%
}
\footnotesize
The p-value of the improvement of \MODELNAME{} over the other methods is marked: * indicating ($p < 0.05$), ** indicating ($p < 0.01$).
\end{table*}

\begin{table*}[htb]
\caption{Ablation study results on SEED, SEED-IV, and DEAP datasets. The asterisks indicate the statistical significance of the improvement of \MODELNAME{} over each variant.}
\label{tab:ablation}
\centering
\resizebox{\textwidth}{!}{%
\begin{tabular}{lcccccc}
\toprule
\multirow{2}{*}{Variant} & \multicolumn{2}{c}{SEED} & \multicolumn{2}{c}{SEED-IV} & \multicolumn{2}{c}{DEAP} \\
\cmidrule(lr){2-3} \cmidrule(lr){4-5} \cmidrule(lr){6-7}
 & Accuracy & F1 & Accuracy & F1 & Accuracy & F1 \\
\midrule
ST & 0.7049 ± 0.1504\textsuperscript{**} & 0.4841 ± 0.1570\textsuperscript{**} & 0.6627 ± 0.1143\textsuperscript{**} & 0.6061 ± 0.1149\textsuperscript{**} & 0.4855 ± 0.0942\textsuperscript{*} & 0.3771 ± 0.0864\textsuperscript{**} \\
TT & 0.7510 ± 0.1318\textsuperscript{**} & 0.5560 ± 0.1578\textsuperscript{**} & 0.7366 ± 0.1011 & 0.6730 ± 0.1146\textsuperscript{*} & 0.4907 ± 0.0845\textsuperscript{*} & 0.3856 ± 0.0769\textsuperscript{*} \\
STT & 0.8031 ± 0.1231 & 0.6322 ± 0.1888\textsuperscript{**} & 0.7386 ± 0.1020 & 0.6920 ± 0.1091 & 0.4991 ± 0.0768\textsuperscript{*} & 0.3954 ± 0.0651\textsuperscript{*} \\
ST-CL & 0.7201 ± 0.1489\textsuperscript{**} & 0.5168 ± 0.1587\textsuperscript{**} & 0.6793 ± 0.1219\textsuperscript{**} & 0.6286 ± 0.1257\textsuperscript{**} & 0.5005 ± 0.0777\textsuperscript{*} & 0.3879 ± 0.0848\textsuperscript{*} \\
TT-CL & 0.7853 ± 0.1185\textsuperscript{**} & 0.6208 ± 0.1551\textsuperscript{**} & 0.7406 ± 0.1036 & 0.6951 ± 0.1202 & 0.5040 ± 0.0754 & 0.4009 ± 0.0681 \\
\midrule
\MODELNAME{} & \textbf{0.8384 ± 0.0695} & \textbf{0.7824 ± 0.1017} & \textbf{0.7652 ± 0.1071} & \textbf{0.7172 ± 0.1200} & \textbf{0.5299 ± 0.0731} & \textbf{0.4226 ± 0.0613} \\
\bottomrule
\end{tabular}%
}
\footnotesize
The p-value of the improvement of \MODELNAME{} over the variant method: * indicating (\(p < 0.05\)), ** indicating (\(p < 0.01\)).
\end{table*}


\subsection{Performance Comparison}
As presented in Table~\ref{tab:performance}, our proposed \MODELNAME{} framework consistently outperforms existing state-of-the-art (SOTA) methods across three benchmark datasets, establishing new performance benchmarks in EEG-based emotion recognition. 

On the SEED dataset, \MODELNAME{} achieves an accuracy of 83.84\% (+2.21\%) and an F1-score of 78.24\% (+4.58\%) over the previous SOTA, FBCCNN. This strong performance extends to SEED-IV, where our model surpasses GCBNet with 76.52\% accuracy (+1.88\%) and a 71.72\% F1-score (+2.87\%). Similarly, on the DEAP dataset, \MODELNAME{} again leads with 52.99\% accuracy (+1.78\%) and a 42.26\% F1-score (+1.67\%) compared to the best prior result from DGCNN.

Furthermore, \MODELNAME{} demonstrates the superiority of its intensity-based curriculum over several alternative CL strategies. It surpasses the self-paced (STT-SPCL) and competence-based (STT-CCL) variants by 2.77\% and 1.39\% in accuracy on SEED and DEAP, respectively. Our model also outperforms the dynamic curriculum (STT-DCL) on SEED-IV by 0.82\% in accuracy and 2.25\% in F1-score. These results validate that our design is crucial for effectively modeling emotional dynamics.

Three key characteristics emerge from the experimental validation: First, the framework demonstrates enhanced robustness with standard deviations lower than previous methods. Second, the performance gains are consistent across datasets, highlighting the method's adaptability. Third, the improvements hold across both accuracy and F1-score metrics, suggesting balanced enhancement in both overall prediction correctness and class-wise performance.

\subsection{Ablation Study}
Table~\ref{tab:ablation} validates the necessity of each component in \MODELNAME{} via three variants: spatial-only (ST), temporal-only (TT), and their CL-enhanced counterparts. The quantitative analysis reveals three critical findings:

(1) Decomposing the model reveals complementary spatiotemporal contributions. On SEED, removing the spatial encoder (TT variant) reduces accuracy by 5.21\%. Conversely, ablating the temporal encoder (ST variant) causes a more severe performance drop of 9.82\% in accuracy. This empirically confirms that temporal dynamics constitute fundamental emotional signatures, while spatial patterns provide discriminative refinement.

(2) The CL extension provides a consistent performance uplift. When applied to the TT baseline, CL achieves a 3.43\% accuracy improvement on SEED (TT-CL), demonstrating its standalone efficacy. Its full integration into the STT architecture yields a 3.53\% accuracy enhancement, underscoring CL's critical role in handling intensity variations through progressive learning.

(3) The analysis reveals an asymmetric dependency on CL. The temporal component exhibits a stronger reliance on CL guidance, with its accuracy improving by 3.43\% (TT vs. TT-CL). In contrast, the spatial component shows a smaller gain of 2.15\% (ST vs. ST-CL). This suggests that temporal dynamics, being more sensitive to emotional intensity, benefit more from the structured learning that CL provides.

These findings establish three conclusions: First, spatiotemporal co-modeling is essential for comprehensive neural representation. Second, curriculum-based learning effectively bridges the gap to real-world emotional variability. Finally, component synergy, not isolated optimization, drives state-of-the-art performance.

\begin{figure}[t]
	\centering 
    \includegraphics[width=0.7\linewidth]{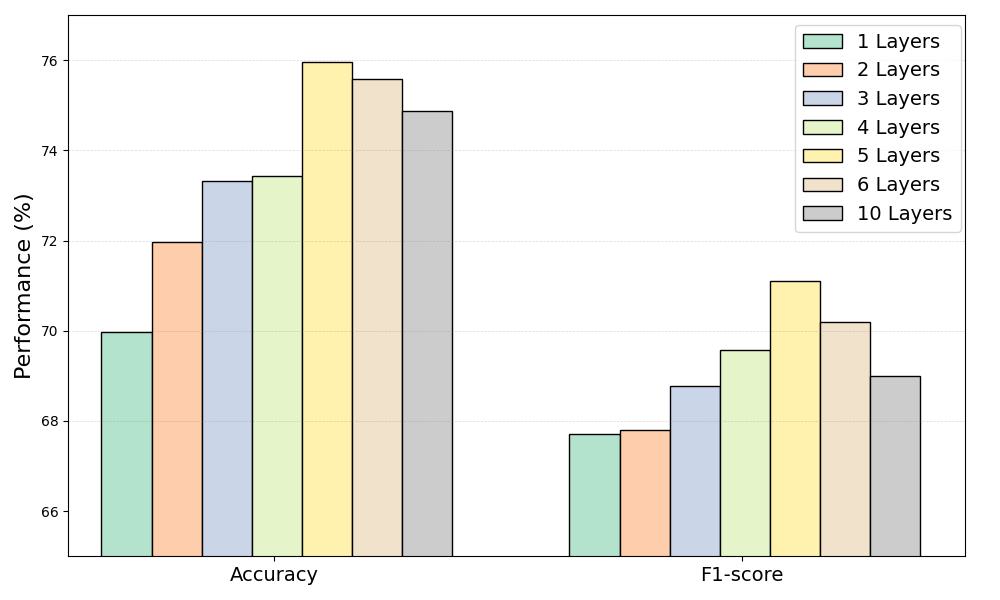} 
    \caption{Hyper-parameter analysis: Encoder layers.}	\label{fig:hyper_params_layers}
\end{figure}

\begin{figure}[t]
	\centering 
    \includegraphics[width=0.7\linewidth]{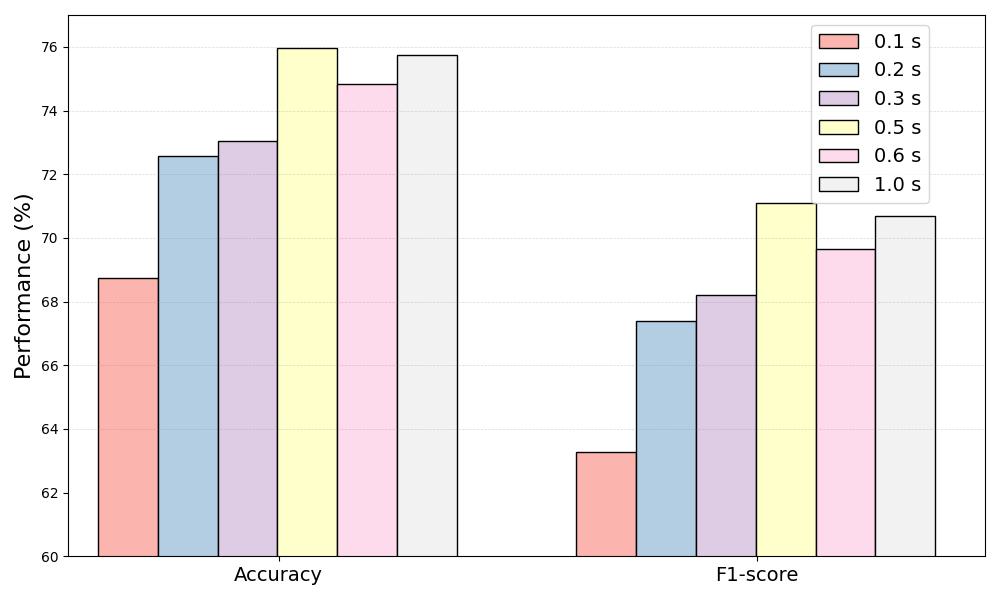} 
    \caption{Hyper-parameter analysis: Time window.}	\label{fig:hyper_params_time_windows}
\end{figure}

\begin{figure}[t]
	\centering 
    \includegraphics[width=0.8\linewidth]{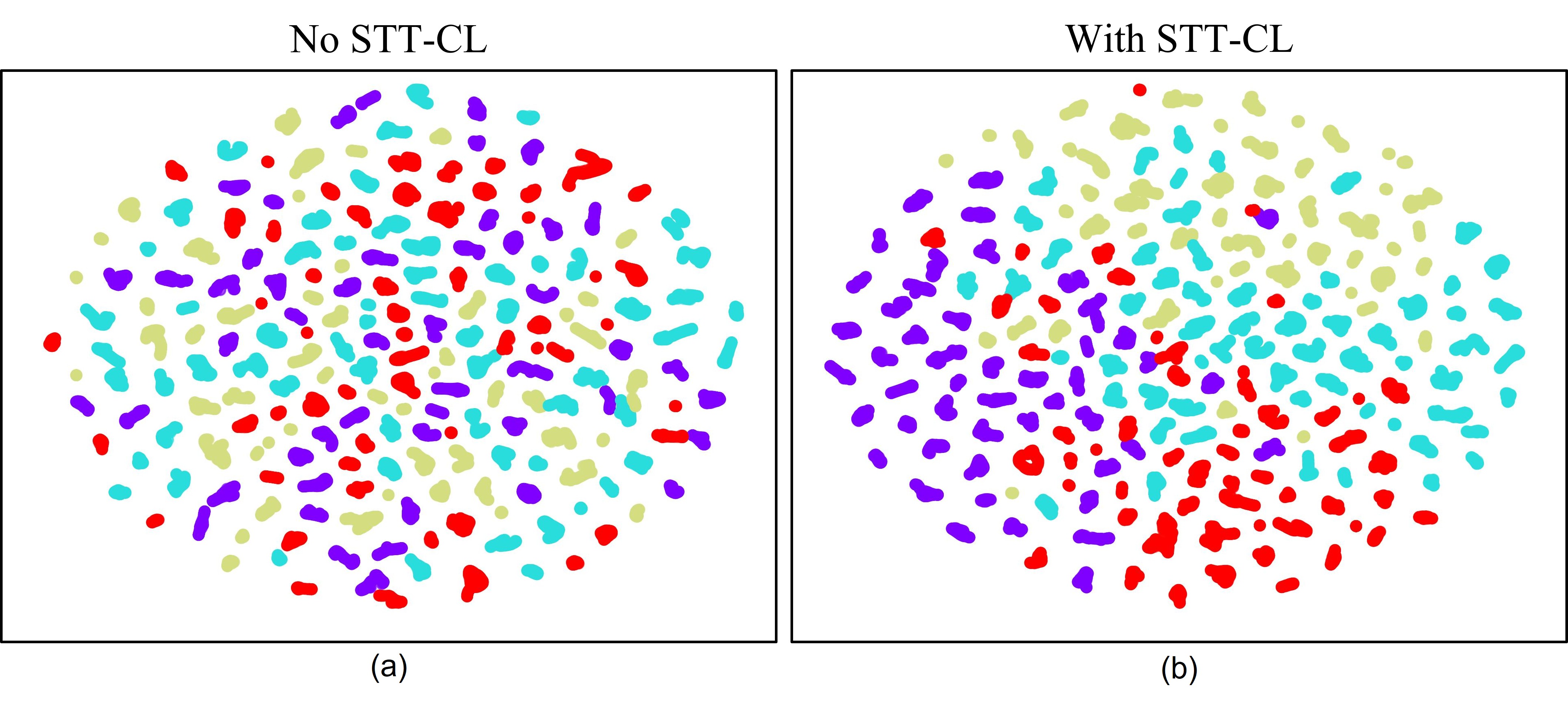} 
    \caption{t-SNE visualization of \MODELNAME{} for feature learning. Different colors represent different categories.}	\label{fig:tSNE}
\end{figure}

\subsection{Hyper-parameter Analysis}

In this section, we analyze the impact of key hyper-parameters, including the number of layers in the Transformer Encoders and the time window size of one segment, on the performance of \MODELNAME{} using the SEED-IV dataset. Following prior work~\cite{zheng2015investigating}, we set the segment length to 3 seconds.  

The results of the encoder layers' analysis are shown in Fig.~\ref{fig:hyper_params_layers}. As shown, increasing the number of layers improves both accuracy and F1-score up to 5 layers, where the highest performance is achieved. Then, increasing the number of layers beyond 5 (to 6 and 10 layers) results in a slight decrease in performance, indicating diminishing returns from additional layers. This suggests that 5 layers strike a good balance between model complexity and performance.

We also analyze the effect of different time window sizes, while keeping the number of layers fixed at 5. As shown in Fig.~\ref{fig:hyper_params_time_windows}, performance improves as the time window increases, with the best results achieved at 0.5 seconds. Smaller time windows (0.1s and 0.2s) result in lower performance, likely due to insufficient temporal resolution.

\subsection{Visualization} 
In this section, We use t-SNE to visualize feature distributions for the SEED-IV dataset with and without the STT-CL framework. In Fig.\ref{fig:tSNE}(a), without STT-CL, features overlap significantly, indicating difficulty in distinguishing subtle emotions. In contrast, Fig.\ref{fig:tSNE}(b) shows that with STT-CL, the feature clusters are more distinct, with reduced intra-class variation and better separation between emotional categories. These results highlight that \MODELNAME{} improves the ability to capture and differentiate subtle emotional states.

\section{Conclusion}
This paper presents \MODELNAME{}, a novel framework addressing two critical challenges in EEG-based emotion recognition: joint modeling of spatiotemporal dependencies and adaptation to emotional intensity variations. To systematically handle intensity fluctuations in real-world scenarios, we propose a curriculum learning paradigm that progressively guides the model from high- to low-intensity emotional states—a first in EEG-based affective computing.

Our framework achieves state-of-the-art performance across three datasets. These improvements stem from two key innovations: (1) The Spatial-Temporal Transformer synergistically models multi-scale temporal dynamics and inter-channel correlations through dual attention mechanisms; (2) The curriculum learning framework dynamically adapts training sequences based on emotional intensity levels, significantly enhancing recognition of subtle emotional patterns. Ablation studies confirm the necessity of both temporal modeling and spatial encoding, while demonstrating the curriculum strategy's critical role. 





\bibliographystyle{IEEEtran}
\balance
\bibliography{reference}

\end{document}